\pdfoutput=1
\documentclass{article} % For LaTeX2e
\usepackage{iclr2025_conference,times}

\usepackage{amsmath,amsfonts,bm}

\def\eqref#1{equation~\ref{#1}}
\def\1{\bm{1}}

\def\vc{{\bm{c}}}

\def\vx{{\bm{x}}}
\def\vy{{\bm{y}}}

\DeclareMathAlphabet{\mathsfit}{\encodingdefault}{\sfdefault}{m}{sl}
\SetMathAlphabet{\mathsfit}{bold}{\encodingdefault}{\sfdefault}{bx}{n}

\usepackage{hyperref}
\usepackage{url}
\usepackage{graphicx}

\newcommand{\ie}{\textit{i.e.}\ }

\title{Measuring Leakage in Concept-Based Methods: An Information Theoretic Approach}

\author{ \thanks{email}, Moritz Vandenhirtz, Sonia Laguna \\[0.5em], Julia E. Vogt
    ETH Zurich \\ Medical Data Science group \\
Department of Computer Science\\
}
\author{Mikael Makonnen \thanks{Correspondence to \texttt{mmakonnen@ethz.ch}} \\
ETH Zurich\\
\And
Moritz Vandenhirtz \\
ETH Zurich\\
\And
Sonia Laguna \\
ETH Zurich\\
\And
Julia E. Vogt \\
ETH Zurich\\
}
\iclrfinalcopy % Uncomment for camera-ready version, but NOT for submission.
\begin{document}

\maketitle

\begin{abstract}
 
Concept Bottleneck Models (CBMs) aim to enhance interpretability by structuring predictions around human-understandable concepts. However, unintended information leakage, where predictive signals bypass the concept bottleneck, compromises their transparency. This paper introduces an information-theoretic measure to quantify leakage in CBMs, capturing the extent to which concept embeddings encode additional, unintended information beyond the specified concepts. We validate the measure through controlled synthetic experiments, demonstrating its effectiveness in detecting leakage trends across various configurations. Our findings highlight that feature and concept dimensionality significantly influence leakage, and that classifier choice impacts measurement stability, with XGBoost emerging as the most reliable estimator. Additionally, preliminary investigations indicate that the measure exhibits the anticipated behavior when applied to soft joint CBMs, suggesting its reliability in leakage quantification beyond fully synthetic settings. While this study rigorously evaluates the measure in controlled synthetic experiments, future work can extend its application to real-world datasets.

\end{abstract}

\section{Introduction}

\paragraph{Concept Bottleneck Models} Concept bottleneck models (CBM) \citep{Koh2020,Lampert2009,Kumar2009} are a simple class of interpretable neural networks typically trained on data points $\left(\vx,\vc,\vy\right)$, comprising the covariates $\vx\in\mathcal{X}$ and targets $\vy\in\mathcal{Y}$ additionally annotated by the concepts $\vc\in\mathcal{C}$. Consider a neural network $f_{\boldsymbol{\theta}}$ parameterised by $\boldsymbol{\theta}$ and a slice $\left\langle g_{\boldsymbol{\psi}}, h_{\boldsymbol{\phi}}\right\rangle$ \citep{Leino2018} s.t.  
\begin{equation}
    f_{\boldsymbol{\theta}}\left(\vx\right)=g_{\boldsymbol{\psi}}\left(h_{\boldsymbol{\phi}}\left(\vx\right)\right)
    \label{eq:slice}
\end{equation}
for all $\vx\in\mathcal{X}$, where $\boldsymbol{\hat{y}}:= f_{\boldsymbol{\theta}}\left(\vx\right)=g_{\boldsymbol{\psi}}\left(h_{\boldsymbol{\phi}}\left(\vx\right)\right)$ denotes the output of the network, that is the predicted target. CBMs enforce a concept bottleneck $\boldsymbol{\hat{c}}:= h_{\boldsymbol{\phi}}(\vx)$, that is the model's final output depends on the covariates $\vx$ solely through the predicted concepts $\boldsymbol{\hat{c}}$.  Thus, in addition to the target prediction loss applied to the final output, $h_{\boldsymbol{\phi}}(\cdot)$ is trained to predict the ground-truth concept values. For more information about how related work tackles this formulation, we refer to Appendix \ref{app:rel_work}.

\paragraph{Interpretability} The interpretability of CBMs is achieved by the set of high-level, human-understandable concepts. Often, these are $C$ binary-valued attributes, i.e. $\mathcal{C}=\left\{0,\,1\right\}^C$ that can be easily detected from the covariates $\vx$ and are predictive of the targets $\vy$. Although CBMs make no assumptions on (anti)causal relationships among $\vx$, $\vc$, and $\vy$, they implicitly assume that concepts $\vc$ are a sufficient statistic \citep{Yeh2020} for predicting $\vy$ based on $\vx$ \citep{Havasi2022, Marcinkevics2023}, i.e. $\vy \perp \vx\,\vert\,\vc$.

\paragraph{Leakage} Leakage is an instance of shortcut learning~\citep{geirhos2020shortcut}. \cite{Margeloiu2021,Mahinpei2021, Havasi2022} show that leakage occurs in cases where the conditional independence assumption does not hold. The distribution of the predicted concept values encodes more information than solely the probability of concept presence. This additional information can then be exploited by the classifier $g_{\boldsymbol{\psi}}\left(\cdot\right)$. This is an issue since the predicted concept values encode information different from the human-understandable concepts, thus, prohibiting the interpretation of the predicted probability as probability of concept presence. \cite{Mahinpei2021} show that even if the predicted concepts are not soft (\ie $\vc\in[0,1]$) but hard (\ie $\vc\in\{0,1\}$), leakage happens, albeit weaker. Therefore, the perception of interpretability for standard CBMs is questionable if $\vy \perp \vx\,\vert\,\vc$ is not fulfilled, which is often the case in real-world problems. Especially, if the concept representation is chosen such that leakage is more likely to appear \citep{espinosa2022concept, ismail2023concept}.

\paragraph{Problem Formulation}

To assess the extent to which the interpretability of the estimated concept embedding is violated, a metric is needed to measure the leakage. This paper introduces an information-theory-inspired measure for leakage in concept-based methods and provides an initial experimental validation.
The anonymized code to reproduce our results can be found \href{https://github.com/MMakonnen/leakage-cbm}{here}.

\section{Methodology and Experimental Setup}

\subsection{Leakage Measure}

First, we introduce the leakage measure that provides the foundation of this paper. Consider a neural network $NN_{\boldsymbol{\theta}}$ parameterized by $\boldsymbol{\theta}$ and a slice $\left\langle g_{\boldsymbol{\psi}}, h_{\boldsymbol{\phi}}\right\rangle$ \citep{Leino2018}, such that  
\begin{equation}
    NN_{\boldsymbol{\theta}}\left(\vx\right) = g_{\boldsymbol{\psi}}\left(h_{\boldsymbol{\phi}}\left(\vx\right)\right).
    \label{eq:slice2}
\end{equation}
For intuition, think of it as a CBM, where $\hat{\vc} = h_{\boldsymbol{\phi}}(\vx)$ is trained to predict concepts from input features. However, this formulation allows for a more general interpretation.

We are interested in quantifying leakage, that is the information contained within the estimated concepts $\hat{\vc}$ that is informative for the label $\vy$ but independent of (or non-informative about) the concepts $\vc$. In information theoretic terms this can be expressed as:
\begin{equation}
    I(\vy;\hat{\vc} \mid \vc) = H(\vy \mid \vc) - H(\vy \mid \hat{\vc}, \vc).
\end{equation}

Estimating $H(\vy \mid \vc)$ and $H(\vy \mid \hat{\vc}, \vc)$ is the primary goal of this paper. A straightforward approximation is given by:
\begin{equation}
    H(\vy \mid \hat{\vc}, \vc) = \mathbb{E}[-\log p(\vy \mid \hat{\vc}, \vc)] \approx -\frac{1}{N} \sum_{i=1}^N \log g_{a, \boldsymbol{\psi}}\left(h_{\boldsymbol{\phi}}\left(\vx_i\right), \vc_i\right)_{y_i},
    \label{eq:leak1}
\end{equation}
\begin{equation}
    H(\vy \mid \vc) = \mathbb{E}[-\log p(\vy \mid \vc)] \approx -\frac{1}{N} \sum_{i=1}^N \log g_{b, \boldsymbol{\psi}}\left(\vc_i\right)_{y_i},
    \label{eq:leak2}
\end{equation}
where $g_{a, \boldsymbol{\psi}}$ and $g_{b, \boldsymbol{\psi}}$ are classifiers trained to predict $\vy$ from $(\hat{\vc},\vc)$ and $\vc$, respectively.

In essence, the idea is that by approximating $I(\vy; \hat{\vc} \mid \vc)$, we obtain an estimate of the leakage in the concept embeddings.

\subsection{Synthetic Data Setup}

% Note here that the estimated concepts (\(\hat{\mathbf{c}}\)) are constructed directly, avoiding the need to train a concept bottleneck model and allowing full control over the leakage. This allows direct testing of whether the measure performs as expected when the leakage is systematically varied. After validating the measure, the same synthetic data (excluding the constructed estimated concepts) can be used to train and test concept-based architectures, allowing controlled leakage comparisons across models.

% In this setup, ground truth concepts are derived from a subset of the feature components, reflecting realistic feature-concept relationships. Estimated concepts incorporate additional, unused feature information to introduce leakage, encoding both the ground truth concepts and extra target-relevant information. Targets are then generated as a function of both the ground truth concepts and the leakage, creating a controlled environment where targets depend on both intended and leakage pathways.

To validate the introduced leakage measure, we first generate a fully synthetic dataset with precisely controlled levels of induced leakage. Since leakage is more pronounced in the soft CBM setting compared to the hard setting \citep{Mahinpei2021}, we perform our experiments in this framework. We then use the synthetic data to assess whether the measure reliably detects the introduced leakage in the controlled setting through simulation experiments. As part of this, we determine suitable parameterizations for the classifiers \( g_{a, \boldsymbol{\psi}} \) and \( g_{b, \boldsymbol{\psi}} \) to ensure accurate estimation.

We have \( n \) observations indexed by \( i = 1, \dots, n \) and \( k \) concepts indexed by \( j = 1, \dots, k \). First, draw the \textbf{features} \(\boldsymbol{x}_i \overset{iid}{\sim} \boldsymbol{X} \in \mathbb{R}^{d}\) where \(\boldsymbol{X} \sim \mathcal{N}\left(\boldsymbol{\mu}_x, \boldsymbol{\Sigma}_x\right)\), with \(\boldsymbol{\mu}_x \in \mathbb{R}^d\) and \(\boldsymbol{\Sigma}_x \in \mathbb{R}^{d \times d}\). Next, the \textbf{binary ground truth concept vector} \(\boldsymbol{c}_i \in \{0,1\}^{k}\) is constructed by sampling each concept from a Bernoulli distribution. This way the inherent uncertainty and noise in the relationship between features and concepts is captured. For each observation, a vector of success probabilities \(\boldsymbol{\pi}_i \in \mathbb{R}^k\), one for each concept, is computed using a function of a subset of the feature information. This approach ensures that the features inform the ground truth concepts while not utilizing all their information, allowing the remaining information to be used later for modeling leakage. Specifically,
\begin{equation}
    c_{ij} \sim \text{Bernoulli}(\pi_{ij}), \quad \boldsymbol{\pi}_i = \sigma\bigl(\boldsymbol{A} \boldsymbol{x}_i + \boldsymbol{\epsilon}_c\bigr),
\end{equation}
where \(\boldsymbol{\epsilon}_c \sim \mathcal{N}\left(0, \boldsymbol{\Sigma}_c\right)\), with \(\boldsymbol{\epsilon}_c \in \mathbb{R}^k\) and \(\boldsymbol{\Sigma}_c \in \mathbb{R}^{k \times k}\). Here, \(\sigma\) denotes the sigmoid activation function, applied element-wise to map the logits to the \([0,1]\) range.

Next, it is important to explain how only a subset of the feature information is used in constructing the success probabilities for the ground truth concepts. This is achieved through the matrix \(\mathbf{A} \in \mathbb{R}^{k \times d}\), which is designed to perform a random projection of the first \(b\) elements of the feature vector \(\boldsymbol{x}_i\) into the \(k\)-dimensional concept space. For this specifically, a random projection, as opposed to another type of projection, is employed to emulate the potentially black-box nature in which concept embeddings are generated by Concept Bottleneck Models. Importantly, the use of a random projection here preserves the relative geometry between observations with high probability, as described by the Johnson-Lindenstrauss lemma. Note that, preferably, the dimensionality of the concept embedding (\(k\)) is chosen to be less than or equal to the number of features being projected (\(b\)); if \(k > b\), the data effectively lies within a \(b\)-dimensional subspace of the \(k\)-dimensional space, meaning that increasing \(k\) beyond \(b\) does not result in any further change to the effective dimensionality. In detail, the matrix \(\mathbf{A}\) is constructed as
\begin{equation}
    \mathbf{A} = \bigl[\, \mathbf{R}_A \ \big| \ \mathbf{0}_{k \times (d - b)}\,\bigr]_{k \times d} \ .
\end{equation}

Here, \(\mathbf{R}_A \in \mathbb{R}^{k \times b}\) is a random projection matrix, and \(\mathbf{0}_{k \times (d - b)}\) is a zero matrix ensuring that the remaining \(d - b\) elements of the feature vector (those beyond the first \(b\), which are projected) do not contribute to the concept generation. The entries of \(\mathbf{R}_A\) are sampled independently from a standard normal distribution, i.e., \(\left(\mathbf{R}_A\right)_{jp} \stackrel{\text{i.i.d.}}{\sim} \mathcal{N}(0,1)\) for \(p = 1, \dots, b\).

Proceeding, the \textbf{estimated concept vector} \(\hat{\boldsymbol{c}}_i \in [0,1]^{k}\) is constructed. While referred to as "estimated concepts," they are not actually estimated but rather constructed in this synthetic data setting to maintain control over the degree of leakage. It is computed as  
\begin{equation}
    \hat{\boldsymbol{c}}_i = \sigma\Bigl( \mathbf{A}\boldsymbol{x}_i + \boldsymbol{l}_i + \boldsymbol{\epsilon}_{\hat{c}} \Bigr),
    \quad \text{with} \quad \boldsymbol{l}_i = \mathbf{B}\boldsymbol{x}_i,
\end{equation}
where \(\boldsymbol{\epsilon}_{\hat{c}} \sim \mathcal{N}\left(0, \boldsymbol{\Sigma}_{\hat{c}} \right)\) (with \(\boldsymbol{\epsilon}_{\hat{c}} \in \mathbb{R}^k\) and \(\boldsymbol{\Sigma}_{\hat{c}} \in \mathbb{R}^{k \times k}\)) introduces noise to model uncertainty, and \(\boldsymbol{l}_i \in \mathbb{R}^{k}\) represents the leakage term. Here, \(\mathbf{B} \in \mathbb{R}^{k \times d}\) is designed to project specific elements of the feature vector \(\boldsymbol{x}_i\) into the concept space, introducing additional information not present in the ground-truth concepts. Similar to the generation of ground-truth concepts, \(\mathbf{B}\) applies a random projection to map features into the concept space. However, \(\mathbf{B}\) specifically projects elements of \(\boldsymbol{x}_i\) from positions \(b+1\) to \(d - l\), effectively utilizing the \(d - b\) features not used in \(\mathbf{A}\) while excluding the last \(l\) features. Precisely, \(\mathbf{B}\) is constructed as  
\begin{equation}
    \mathbf{B} = \bigl[ \, \mathbf{0}_{k \times b} \ \big| \ \mathbf{R}_B \ \big| \ \mathbf{0}_{k \times l} \, \bigr]_{k \times d} \ .
\end{equation}

Here, \(\mathbf{0}_{k \times b}\) and \(\mathbf{0}_{k \times l}\) are zero matrices, ensuring that the first \(b\) and last \(l\) elements of the feature vector are excluded. The matrix \(\mathbf{R}_B \in \mathbb{R}^{k \times (d - b - l)}\) is a random projection matrix with entries \(\left( \mathbf{R}_B \right)_{jq} \stackrel{\text{i.i.d.}}{\sim} \mathcal{N}(0,1)\) for \(q = 1, \dots, d - b - l\).

Lastly, the \textbf{target variable} \(\boldsymbol{y}_i \in \{1, \dots, J\}\) is constructed, where \( J \in \mathbb{N} \). Operating in a multiclass setting and to introduce randomness, analogous to the earlier argument for sampling from a Bernoulli distribution rather than thresholding, the target \(y_i\) is sampled as  
\begin{equation}
    y_i \sim \text{Categorical}\bigl(\boldsymbol{p}_i\bigr), \quad \boldsymbol{p}_i = \operatorname{softmax}\Bigl(f\bigl(\boldsymbol{c}_i, \boldsymbol{l}_i\bigr) + \boldsymbol{\epsilon}_y\Bigr),
\end{equation}
where the probability vector \(\boldsymbol{p}_i \in \mathbb{R}^J\) is computed using a nonlinear function \(f: \mathbb{R}^k \times \mathbb{R}^k \rightarrow \mathbb{R}^J\) that combines the ground-truth concepts \(\boldsymbol{c}_i\) and the leakage term \(\boldsymbol{l}_i\). The noise vector \(\boldsymbol{\epsilon}_y \sim \mathcal{N}(0, \boldsymbol{\Sigma}_y)\) introduces additional randomness, where \(\boldsymbol{\epsilon}_y \in \mathbb{R}^J\) and \(\boldsymbol{\Sigma}_y \in \mathbb{R}^{J \times J}\). Here we parametrize \(f\) as a simple Multi-Layer Perceptron (MLP) with one hidden layer, defined as  
\begin{equation}
    f\bigl(\boldsymbol{c}_i, \boldsymbol{l}_i\bigr) = \mathbf{W}_2\,\phi\Bigl(\mathbf{W}_1 \begin{bmatrix} \boldsymbol{c}_i \\ \boldsymbol{l}_i \end{bmatrix} + \boldsymbol{b}_1\Bigr) + \boldsymbol{b}_2,
\end{equation}
where the input consists of the concatenated ground-truth concepts and leakage term, \(\begin{bmatrix} \boldsymbol{c}_i \\ \boldsymbol{l}_i \end{bmatrix} \in \mathbb{R}^{2k}\). The weight matrix \(\mathbf{W}_1 \in \mathbb{R}^{h \times 2k}\) has entries sampled from \(\mathcal{N}(0,1)\), and the bias vector \(\boldsymbol{b}_1 \in \mathbb{R}^h\) is initialized to zeros. Similarly, the weight matrix \(\mathbf{W}_2 \in \mathbb{R}^{J \times h}\) is sampled from \(\mathcal{N}(0,1)\), and the bias vector \(\boldsymbol{b}_2 \in \mathbb{R}^J\) is initialized to zeros. The activation function \(\phi(z) = \max(0, z)\) is the ReLU function and is applied element-wise.

By constructing the target labels \(y_i\) with this nonlinear function that integrates both ground truth concepts and leakage information, we ensure that the ground truth concepts are informative for predicting the target, while the leakage provides additional information to improve prediction accuracy. Note that this setup ensures the target implicitly depends on the original features through both the ground truth concepts and the leakage term.

To conclude, there are two avenues to control leakage: via \(b\) (by choosing the number of elements from the feature vector that enter as information into the ground truth concepts, and thus do not contribute to the leakage) and via \(l\) (which provides a finer control over how much of the remaining information from the feature vector contributes to the leakage). Note that \(b, d, h, k, l, n \in \mathbb{N}\).

\subsection{Experimental Validation of Leakage Measure on Fully Synthetic Data}
\label{exp_val}

Having established a leakage measure and a synthetic data generation process with controlled leakage in concept embeddings, the next step is to rigorously validate the measure. To this end, we designed a simulation experiment to assess its ability to detect varying leakage levels, evaluate its robustness, and determine suitable parametrizations for the classifiers \( g_{a, \boldsymbol{\psi}} \) and \( g_{b, \boldsymbol{\psi}} \).

\paragraph{Classifier selection} A critical aspect of our experimental setup is selecting the most suitable parametrizations of \( g_{a, \boldsymbol{\psi}} \) and \( g_{b, \boldsymbol{\psi}} \) from \ref{eq:leak1} and \ref{eq:leak2}. For this we compare: \textbf{Simple Multilayer Perceptron (MLP)} (a simple neural network with one hidden fully connected layer), \textbf{Random Forest} (an ensemble learning method effective for handling tabular data with diverse feature interactions), and \textbf{XGBoost} (a gradient boosting framework known for its high performance and scalability on structured data). Each classifier is implemented using default architectures and parameter settings unless specified differently.

\paragraph{Calibration}  
Accurate entropy estimation in \ref{eq:leak1} and \ref{eq:leak2} is crucial for the validity of the proposed leakage measure. Since entropy is derived from the negative log-likelihoods of predicted probabilities, calibrating classifier predictions, i.e. ensuring that a model’s confidence aligns with its actual correctness, is essential to prevent biased or unreliable entropy values. To achieve this, we apply temperature scaling, a simple yet effective technique that adjusts the model’s logits without retraining, preserving classification accuracy while improving probability calibration. This results in more accurate entropy estimates and, consequently, more reliable leakage measurements.

\paragraph{Experimental Configurations \& Evaluation}  
To evaluate the leakage measure comprehensively, we explore various experimental configurations by adjusting key parameters. The challenge is balancing computational constraints with testing a sufficiently diverse set of configurations to ensure robustness across different scenarios. The chosen settings assess the measure's reliability across dataset sizes, concept complexities, and feature dimensionalities while maintaining feasibility.  

\subparagraph{Fixed Parameters}  
The following parameters remain constant across all configurations:  
\begin{itemize}
    \item \textbf{Number of Target Classes (\( J \))}: Set to 5, representing a typical multiclass classification task.
    \item \textbf{Number of Simulation Runs}: Each configuration is averaged over 5 runs for statistical reliability.
    \item \textbf{Train-Validation-Test Split}: Fixed at 70\% training, 15\% validation, and 15\% testing.
    \item \textbf{Neural Network Settings}: 20 training epochs, batch size of 64.
\end{itemize}  

\subparagraph{Varied Parameters}  
These parameters change across different experimental configurations:  
\begin{itemize}
    \item \textbf{Dataset Size (\( n \))}: 500, 2,000, and 10,000 observations to simulate small to large datasets.
    \item \textbf{Noise Levels}: Diagonal noise constants of 0.5 and 2, representing low and moderate noise levels. Initial experiments focus on varying diagonal variance in covariance matrices.
    \item \textbf{Feature Dimensionality (\( d \))}: 500 and 2,500 features, covering moderate to high-dimensional settings from tabular to image-like data.
    \item \textbf{Number of Concepts (\( k \))}: 50 and 200 concepts to represent low- and high-complexity scenarios, ensuring \( k < d \) given the concept bottleneck structure.
    \item \textbf{Classifier Types}: MLP, Random Forest, and XGBoost to test different modeling approaches.
\end{itemize}  

Each experimental configuration is run across 30 predefined leakage levels (\( k < b < d-k-l \)), with \( l \) initially set to zero for a simpler leakage control. The estimated leakage is averaged over 5 simulation runs per configuration.

In summary, the primary objectives of this experimental validation are to confirm the \textbf{Validity} of the leakage measure across different settings, ensure its \textbf{Robustness} under diverse conditions (varying dataset sizes, noise levels, feature dimensions, and concept complexities), and identify the classifiers \( g_{a, \boldsymbol{\psi}} \) and \( g_{b, \boldsymbol{\psi}} \) that provide the most consistent and robust leakage estimates.

\section{Results}

\begin{figure}[h]
\vspace{1cm}
    \centering
    \includegraphics[width=1.0\textwidth]{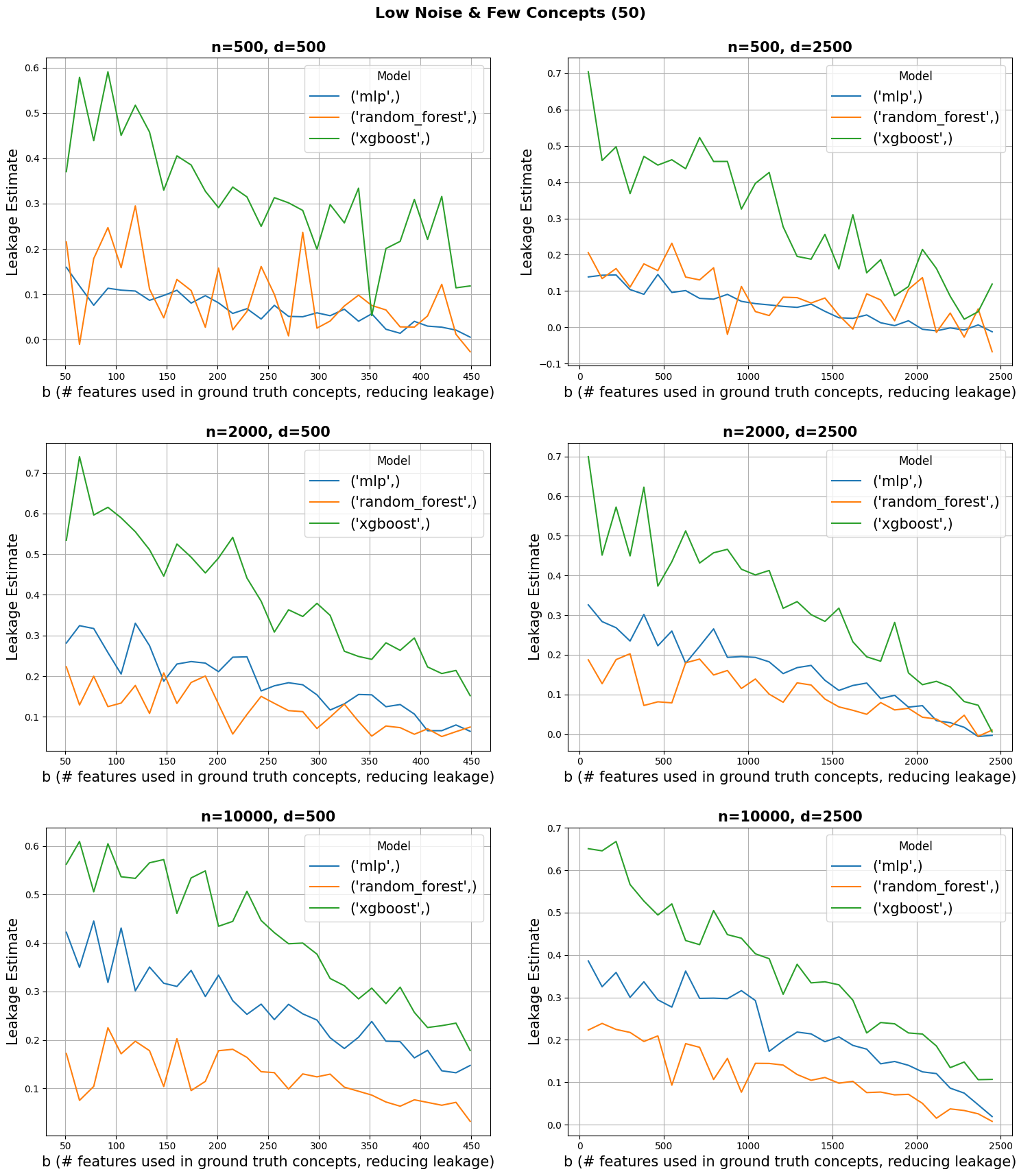}
    \caption{Testing leakage measure on fully synthetic data for low noise and few concepts.}
    \label{fig:illustration1}
%\vspace{2cm}
\end{figure}

The following observations and discussion relate to Figure \ref{fig:illustration1} and Figures \ref{fig:illustration2}, \ref{fig:illustration3}, and \ref{fig:illustration4} from \ref{app:ill}. These figures present the results of testing the leakage measure through simulations across various configurations. The plots display the average estimated leakage over multiple simulation runs, as introduced in Section \ref{exp_val} . Four sets of plots correspond to different noise-concept regimes and the configurations within them.

\paragraph{Theoretical Expectations}

From a theoretical perspective, if the measure behaves as expected, increasing \( b \) should reduce the amount of feature information captured outside the ground-truth concepts, thereby decreasing leakage. Consequently, we expect the leakage measure to reflect this by decreasing accordingly. Therefore, we aim to identify classifier parameterizations that most consistently and robustly satisfy this expectation across different settings.

\paragraph{General Trends}

Figure \ref{fig:illustration1} shows a consistent downward trend in estimated leakage as \( b \) increases across nearly all experimental configurations and classifier parameterizations, aligning with theoretical expectations. However, the magnitude and consistency of this decrease vary significantly depending on the classifier and specific experimental setup. A similar trend of decreasing leakage estimates with increasing \( b \) is observed in the figures \ref{fig:illustration2}, \ref{fig:illustration3}, and \ref{fig:illustration4} from the appendix, which correspond to different noise-concept regimes. However, volatility varies across configurations and parameterizations, making the trend less stable in some cases. In addition to expected patterns — such as estimates stabilizing with larger dataset sizes and the leakage measure being more stable and closer to expectations in lower-noise settings - the following key observations emerge.

\paragraph{Feature Vector Dimensionality}

Across all four regimes, feature vector dimensionality significantly impacts how closely the measure aligns with expectations. In all settings, when comparing the low-dimensional (\( d = 500 \)) to the high-dimensional (\( d = 2500 \)) case while keeping other factors constant, the measure tends to follow expectations more closely in high-dimensional settings. Specifically, it exhibits a clear negative trend as \( b \) increases and does so in a relatively stable manner.  Even when \( d \ll n \), the measure remains effective, though slightly more volatile than in high-dimensional cases. However, in low-dimensional settings, performance improves with fewer concepts and/or lower noise.

\paragraph{Concept Vector Complexity}

The number of concepts (\( k \)) significantly affects leakage estimates, as seen when comparing the results in Figure \ref{fig:illustration1} and Figure \ref{fig:illustration2}, as well as Figure \ref{fig:illustration3} and Figure \ref{fig:illustration4}. When fewer concepts (\( k = 50 \)) are used, leakage estimates are generally higher. A possible explanation is that the limited bottleneck capacity increases the potential for leakage, making accurate entropy estimation more challenging. Conversely, with more concepts (\( k = 200 \)), leakage estimates decrease more consistently as \( b \) increases. This suggests that higher concept dimensionality enhances the model's ability to encode ground-truth information more faithfully, thereby reducing leakage. 

% However, the added complexity also introduces some noise into the estimates, particularly for MLP and Random Forest.

\paragraph{Negative Leakage Results}

A few cases in the experiments show negative leakage estimates, which theoretically should not occur since leakage represents nonnegative information content. However, these values likely arise from implementation-related limitations and increased system noise that affect the leakage estimator. Notably, all instances occur in the smallest dataset setting (\( d = 500 \)), where a limited number of simulation runs amplifies randomness and perturbations. Negative leakage values are more frequent in specific classifier configurations, while some methods, like XGBoost, show greater robustness. Crucially, these occurrences do not undermine the validity of the measure. Instead, they serve as engineering indicators, while the overall trend remains meaningful, correctly capturing the expected signal.

\paragraph{Classifier Performance}

XGBoost is the most reliable classifier for the leakage measure, consistently showing a clear downward trend in leakage as \( b \) increases. It remains stable, aligns well with expectations, and rarely produces negative leakage estimates. While it experiences some fluctuations in high-noise and many-concept settings, its trends are clearer than those of other classifiers.  

MLP performs reasonably well, particularly in low-noise settings, where it shows clear trends with moderate stability. However, in high-noise and complex scenarios, it becomes more variable and occasionally produces negative leakage estimates, likely due to its limited capacity to model complex relationships due to having only one hidden layer.  

Random Forest is the least robust, frequently generating negative leakage estimates and generally producing small magnitudes with no meaningful variation in the measures across different leakage values, especially in high-noise settings and with small datasets. Even under low-noise conditions, it remains highly variable and less reliable than XGBoost. Though it benefits from larger datasets, its weaknesses in probability estimation and tendency to overfit make it less suitable for this task.

\begin{figure}[t]
    \centering
    \includegraphics[width=0.95\textwidth]{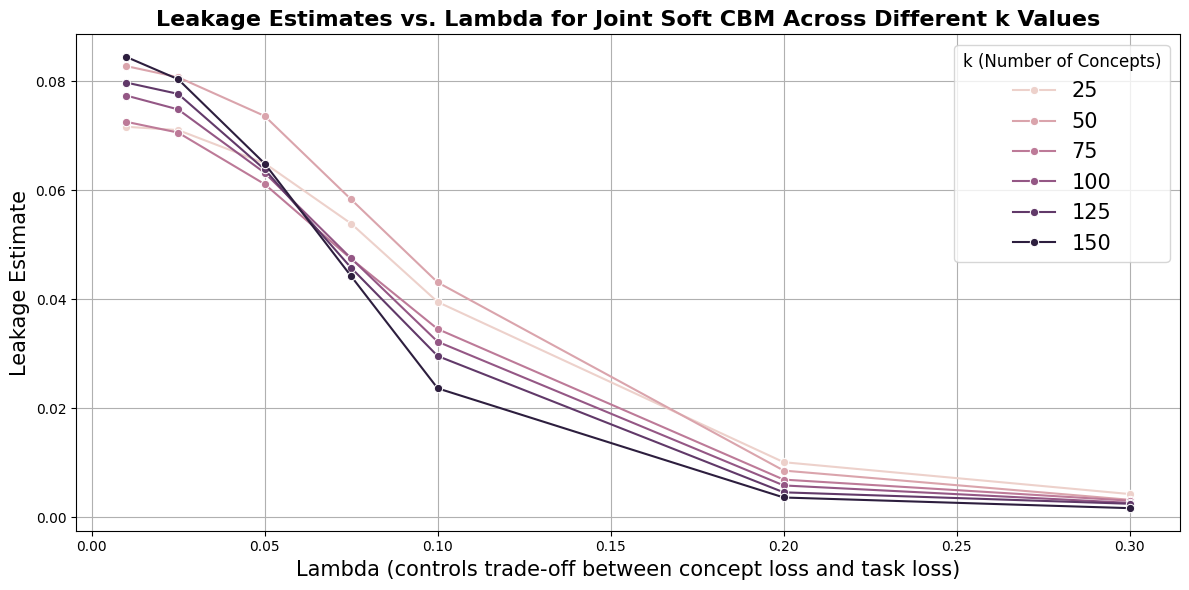}
    \caption{Illustration of relationship between leakage estimates and the regularization parameter $\lambda$ in joint soft CBM for varying numbers of concepts.}
    \label{fig:ill_leak_lambda}
\end{figure}
\section{Conclusion}

The results confirm that the proposed leakage measure effectively captures leakage trends, with a consistent decrease as \( b \) increases across different configurations. XGBoost proves to be the most reliable classifier parametrization, maintaining stability and alignment with theoretical expectations across varying noise levels, dataset sizes, and concept complexities. In contrast, MLP and Random Forest exhibit greater variability, particularly in high-noise settings and with limited data. The findings emphasize the importance of classifier selection, noise control, and dataset size in obtaining reliable leakage estimates. Higher-dimensional feature and concept spaces improve measure stability, reinforcing the theoretical link between bottleneck capacity and leakage. Overall, the measure demonstrates robustness, providing a potential tool for assessing leakage in concept-based models.

Having established the potential value of the proposed leakage measure, we now discuss its limitations. The study relies entirely on synthetic data, which, while a valid starting point for a new measure, limits its generalizability to real-world applications. Further sensitivity analyses and ablation studies are needed to assess more complex noise regimes, diverse configurations, and finer-grained leakage evaluations. Additionally, only a basic calibration technique was used, underscoring the need for more robust calibration methods to ensure reliability.

These limitations naturally suggest future research directions. A logical next step is to transition from a fully synthetic setting to a more realistic, partially synthetic setup, gradually assessing the measure’s applicability in practical scenarios. Specifically, while synthetic features, targets, and ground-truth concepts would still be used, the predicted concepts would now be obtained from a concept-based model, such as the joint soft CBM, trained on these synthetic components. This approach preserves control over the data-generating process while enabling an evaluation of leakage in the models predicted concept embeddings.

A particularly interesting aspect to investigate in this setting is how leakage varies with the regularization parameter \(\lambda\), which controls the trade-off between concept loss and task loss \citep{Koh2020}. Higher values of \(\lambda\) emphasize concept learning, whereas lower values prioritize task performance. We hypothesize that as \(\lambda\) decreases, leakage increases due to a shift in the model’s objective toward optimizing task accuracy, potentially encouraging shortcut learning and unintended information flow. Ideally the leakage measure can pick up on this.

To explore this, we conducted a preliminary experiment in a high-leakage setting (\(n=10,000\), \(d=1,000\), \(J=5\), \(b=160\), \(l=0\)), where we computed leakage across different regularization parameter values and concept dimensions. The results, shown in Figure \ref{fig:ill_leak_lambda}, indicate a consistent trend: as \(\lambda\) increases, leakage decreases across various concept dimensions. Notably, the overall magnitude of leakage in this setup is lower than in the fully synthetic case, suggesting that while this remains a preliminary investigation, it provides empirical support for the measure’s validity beyond a fully synthetic setting. 
Lastly, we provide other exciting future research direction in App. \ref{app:future_work}.

\section*{Acknowledgments}
MV and SL is supported by the Swiss State
Secretariat for Education, Research, and Innovation (SERI) under contract number MB22.00047.

\newpage

\bibliography{iclr2025_conference}
\bibliographystyle{iclr2025_conference}

\appendix
\section{Appendix}

\subsection{Related Work}
\label{app:rel_work}
To measure leakage, \cite{zarlenga2023towards} propose metrics that estimate the degree of excessive information with respect to other concepts, which they call impurity. To resolve leakage, \cite{Margeloiu2021} recommend using the independent training procedure with hard concepts, where the concept encoder and classification head are trained entirely independently, with ground-truth concepts provided as inputs to the classification head during training. However, this approach reduces performance since the encoder and predictor head cannot communicate during the training process. Thus, \cite{Havasi2022} propose to include a hard side-channel, in which the additional information can be learned explicitly, as well as an autoregressive structure over the hard concept predictions, such that their correlations can be captured. At intervention time, they use importance-weighted MCMC sampling to implicitly learn the effect of a concept intervention on the other concepts. \cite{vandenhirtzstochastic} build upon the idea of modeling hard concepts and learn a logit-normal distribution to avoid the slow autoregressive structure.

\subsection{Additional Future Directions}
\label{app:future_work}

\paragraph{Exploring Alternative Classifiers and Methods}

Investigating alternative classifiers, such as probabilistic neural networks or ensemble methods, could further improve the robustness of CBMs against leakage. Additionally, replacing the embedding \( \mathbf{z} \) with \( g_{\boldsymbol{\psi}}(\mathbf{z}) \) may help capture leakage within the embeddings more effectively.

% \paragraph{Applying to Real-World Datasets}

% Applying the leakage measure to real-world datasets (e.g., CUB, CelebA, CIFAR-10) is essential to validate its utility beyond synthetic environments and to broaden its applicability. This transition will likely require adaptations and tuning of metric approximations and estimators to handle the complexities of real-world data.

\paragraph{Leakage as a Regularizer}

Exploring the use of the leakage measure as a regularizer during training could provide a method for mitigating leakage in jointly trained CBMs. This approach could help enforce tighter control over unintended information flow.

\paragraph{Normalizing Mutual Information}

To bound the leakage metric, mutual information \( I(\mathbf{z}; \mathbf{y} \mid \mathbf{c}) \) could be normalized by the maximum possible information \( I(\mathbf{x}; \mathbf{y} \mid \mathbf{c}) \). This involves training an additional approximator to estimate \( H(\mathbf{y} \mid \mathbf{x}, \mathbf{c}) \), resulting in a normalized leakage metric:
\[
\frac{ I(\mathbf{z}; \mathbf{y} \mid \mathbf{c}) }{ I(\mathbf{x}; \mathbf{y} \mid \mathbf{c}) }
\]

\paragraph{Estimator Integration}

Investigating whether estimators \( g_a \) and \( g_b \) should remain separate or be combined into a single estimator could improve the accuracy and consistency of leakage measurements. Ensuring that \( H(\mathbf{y} \mid \mathbf{c}) \geq H(\mathbf{y} \mid \mathbf{z}, \mathbf{c}) \) is crucial for maintaining the validity of the leakage measure.

\newpage
\subsection{Illustrations}
\label{app:ill}

\begin{figure}[h]
    \centering
    \includegraphics[width=1.05\textwidth]{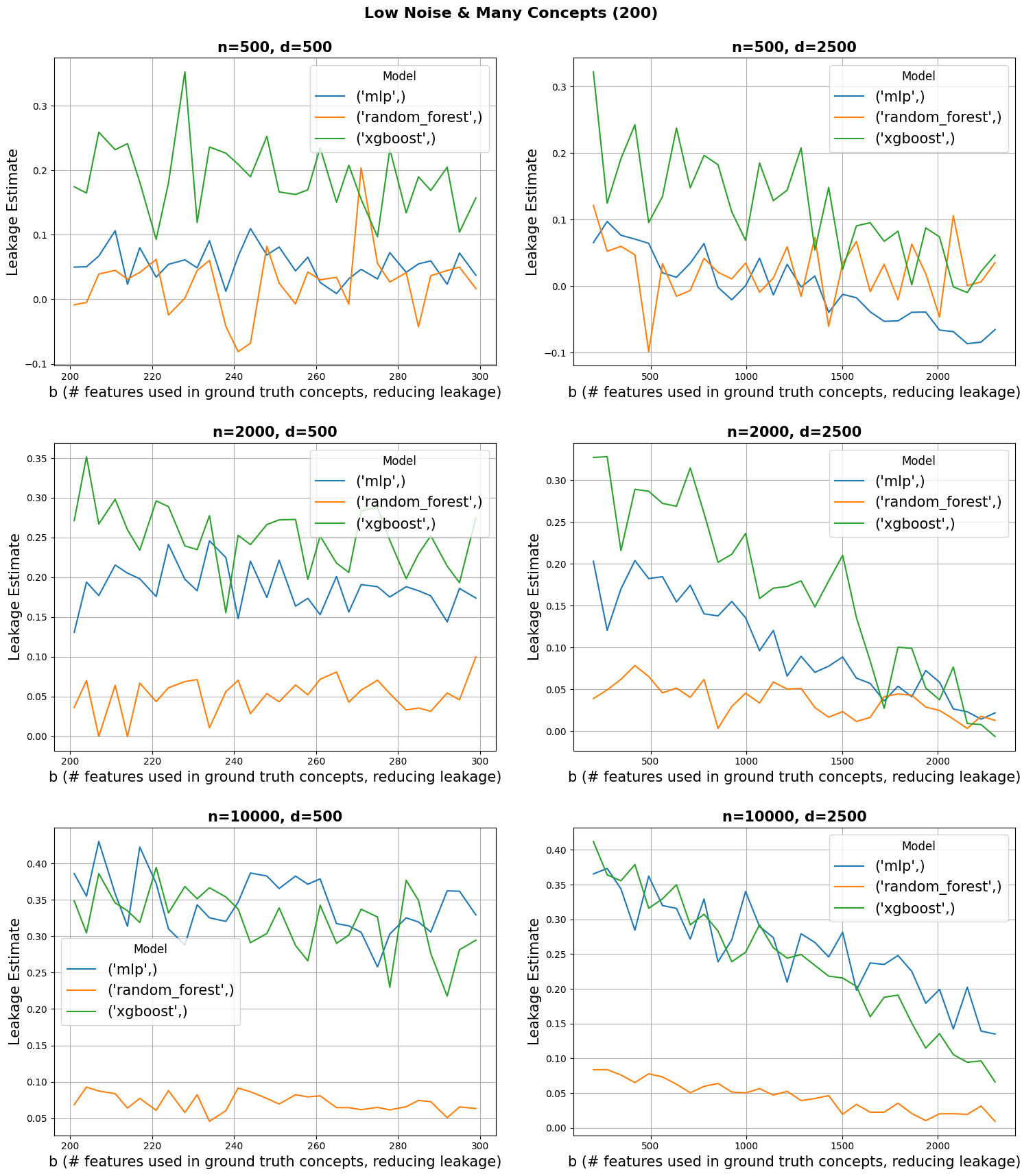}
    \caption{Testing leakage measure on fully synthetic data for low noise and many concepts.}
    \label{fig:illustration2}
\end{figure}

\begin{figure}[h]
    \centering
    \includegraphics[width=1.05\textwidth]{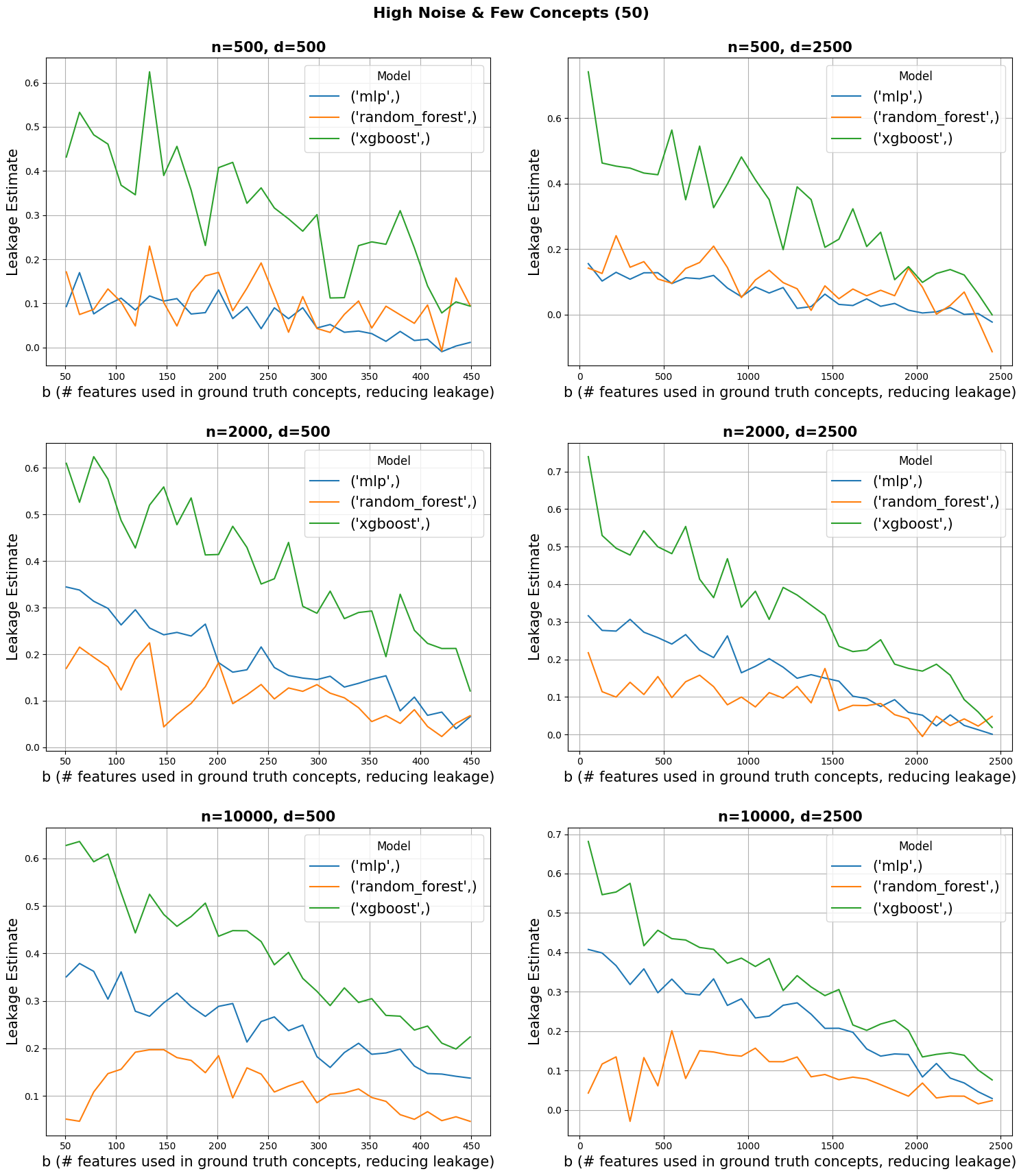}
    \caption{Testing leakage measure on fully synthetic data for high noise and few concepts.}
    \label{fig:illustration3}
\end{figure}

\begin{figure}[h]
    \centering
    \includegraphics[width=1.05\textwidth]{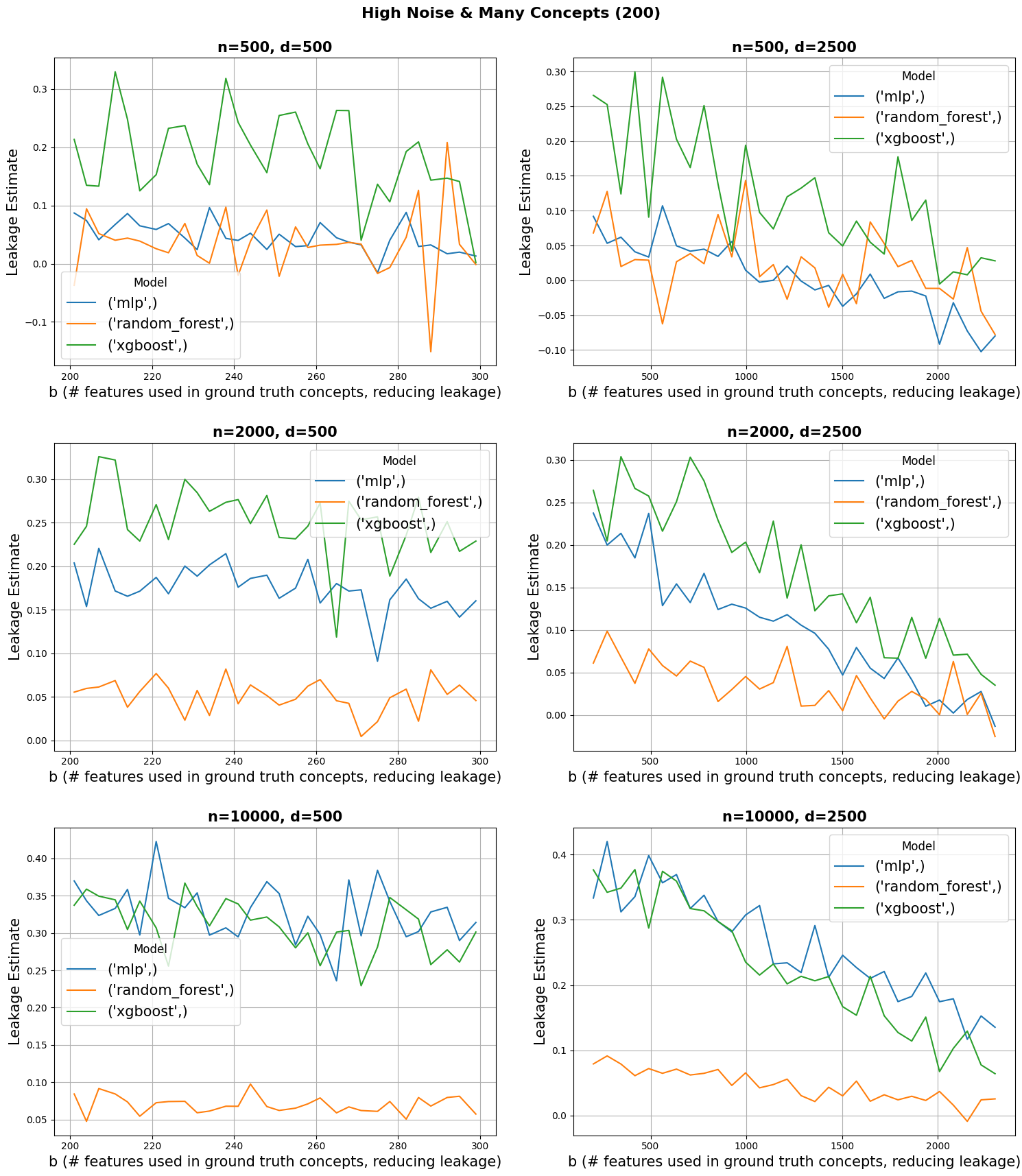}
    \caption{Testing leakage measure on fully synthetic data for high noise and many concepts.}
    \label{fig:illustration4}
\end{figure}

% \clearpage  % Ensures all figures are placed before moving on

\end{document}